\documentclass[letterpaper]{article} 
\usepackage{aaai25}  
\usepackage{times}  
\usepackage{helvet}  
\usepackage{courier}  
\usepackage[hyphens]{url}  
\usepackage{graphicx} 
\usepackage{natbib}  
\usepackage{caption} 
\usepackage{algorithm}
\usepackage{algorithmic}

\usepackage{newfloat}
\usepackage{listings}
\DeclareCaptionStyle{ruled}{labelfont=normalfont,labelsep=colon,strut=off} 
\floatstyle{ruled}
\newfloat{listing}{tb}{lst}{}
\floatname{listing}{Listing}

\makeatletter

\newcommand{\Rmnum}[1]{\expandafter\@slowromancap\romannumeral #1@}
\makeatother
\usepackage[utf8]{inputenc} 
\usepackage{booktabs}       
\usepackage{amsfonts}       
\usepackage{nicefrac}       
\usepackage{microtype}      
\usepackage{xcolor}         
\usepackage{subfigure}
\usepackage{soul}
\usepackage{makecell}
\usepackage{multirow}
\usepackage{threeparttable}
\usepackage{arydshln}
\usepackage{lipsum}

\usepackage{amsmath}
\usepackage{amssymb}
\usepackage{mathtools}
\usepackage{amsthm}

\usepackage[capitalize,noabbrev]{cleveref}

\title{Order of Compression: A Systematic and Optimal Sequence to Combinationally Compress CNN}
\author {
    Yingtao Shen\textsuperscript{\rm 1},
    Minqing Sun\textsuperscript{\rm 1},
    Jianzhe Lin\textsuperscript{\rm 2},
    Jie Zhao\textsuperscript{\rm 2},
    An Zou\textsuperscript{\rm 1},
}
\affiliations {
    \textsuperscript{\rm 1}Shanghai Jiao Tong University\\
    \textsuperscript{\rm 2}Microsoft Corporation\\
    doctorcoal@sjtu.edu.cn, sunminqing@sjtu.edu.cn, jianzhelin@microsoft.com, zhaojie@microsoft.com, an.zou@sjtu.edu.cn
}
\begin{document}
\maketitle
\begin{abstract}
Model compression has gained significant popularity as a means to alleviate the computational and memory demands of machine learning models. Each compression technique leverages unique features to reduce the size of neural networks. Although intuitively combining different techniques may enhance compression effectiveness, we find that the order in which they are combined significantly influences performance. To identify the optimal sequence for compressing neural networks, we propose the Order of Compression, a systematic and optimal sequence to apply multiple compression techniques in the most effective order. We start by building the foundations of the orders between any two compression approaches and then demonstrate inserting additional compression between any two compressions will not break the order of the two compression approaches. Based on the foundations, an optimal order is obtained with topological sorting. Validated on image-based regression and classification networks across different datasets, our proposed Order of Compression significantly reduces computational costs by up to 859$\times$ on ResNet34, with negligible accuracy loss (-0.09\% for CIFAR10) compared to the baseline model. We believe our simple yet effective exploration of the order of compression will shed light on the practice of model compression. 

\end{abstract}
\section{Introduction}
\label{intro}
Deep Learning models have gained considerable attention due to their wide applicability across diverse domains. However, deploying deep learning models on resource-constrained systems such as mobile and embedded systems is challenging because of their high computational and energy costs \cite{ke2018nnest}. To address this challenge, a variety of compression techniques have been developed to reduce the computational cost of these intricate models. 

These approaches operate at different granularities or stages, either offline or dynamically at runtime. For example, before the network is executed, knowledge distillation compresses the architecture of the neural network, pruning removes neurons within layers, and quantization reduces the hardware bits used in the computation of each neuron \cite{mishra2020survey, neill2020overview}. Meanwhile, dynamic inference techniques such as early exit or layer skipping actively compress the network structure, adapting to each input \cite{han2021dynamic} at the inference runtime. 
Several studies have used multiple compression techniques, particularly for CNN-based models. For example, \cite{qi2021learning} combine pruning and quantization to compress CNNs, achieving approximately a 50\% reduction in FLOPs with only 0.15\%-0.37\% accuracy drops. \cite{li2023predictive} integrate early exit with quantization on CNNs, which not only avoids floating point computation but also leads to over 50\% reduction in FLOPs with 1\%-3\% accuracy drops. More complex combinations include Deep Compression \cite{han2015deep}, employing pruning, trained quantization, and Huffman coding to minimize the storage of network parameters and the energy cost in the execution of neural networks; and the Deep Hybrid Compression Network \cite{zhao2023deep}, using quantization, relaxed weight pruning, and knowledge distillation to overcome uniform quantization limitations. Recent state-of-the-art techniques such as APQ \cite{wang2020apq} started to design a Pruning and Quantization aware NAS (Network Architecture Search) method, which requires recognition and modification of specific network architecture to achieve a high compression ratio.

However, most of the existing works are focused on proposing a possible combinational method, ignoring the optimal sequence in applying these techniques. Consequently, a practical yet significant question arises: \textit{Is it necessary to use multiple compression approaches; if so, what is the optimal sequence to apply different compression approaches?} 

\begin{figure}[t]
\centering
\includegraphics[width=0.36\textwidth]{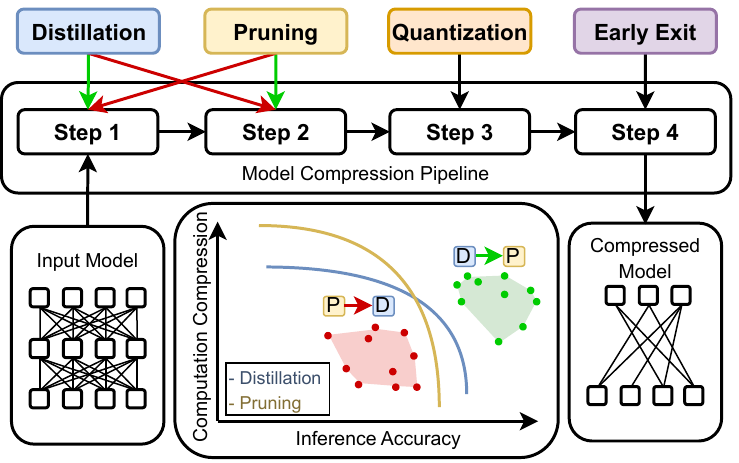}
\caption{The Order of Compression Matters: Distillation then pruning will outperforms pruning then distillation or single pruning or single distillation.
}
\label{fig:intro}
\vspace{-2mm}
\end{figure}
\textbf{Contribution. }In this work, we introduce the Order of Compression, a roadmap to explore the optimal order for combinational compression by treating each compression method as a standard building block, as illustrated in Fig.~\ref{fig:intro}. This modular design enables greater flexibility and adaptability in creating compression strategies customized to specific network architectures and performance requirements. The proposed systematic road map finds the best empirical combinational sequence, focusing on determining in what order the compression method(s) should be chained together to achieve optimal performance.

To explore the order of compression, we start with examining the interaction and sequence of any two compression methods and comparing the pairwise order. This examination yields valuable insights that posit when pairwise orders are combined, they form a directed acyclic graph containing a single choice of topological sorting. This hypothesis is further corroborated through our extensive experiments, demonstrating that the order remains consistent as we insert other methods in between. Our final solution leverages topological sorting, establishing multiple compression methods within an optimal combinational sequence. Experiments demonstrate that applying multiple compressions with the proposed optimal sequence can reduce the model size by $25-1000\times$ with an ignorable accuracy loss. 

The contributions and insights of this paper are as follows:
\begin{itemize}
\item{Combining compression methods may introduce additional benefits. Despite each method's performance varying greatly, they may complement each other.}

\item{Order matters and the optimal order between two methods is likely to be consistent if multiple compression methods are used together. This suggests a systematic way to explore the optimal sequence in compressing networks.}

\item{The optimal sequence to apply multiple compression approaches is derived with topological sorting, which is from static to dynamic and granularity large to small.}

\item Extensive experiments with the mainstream models and tasks demonstrate that multiple compressions with the proposed optimal sequence can reduce the model size up to 25$\times$-1000$\times$ with an ignorable accuracy loss.

\end{itemize}


\section{Design Overview}
\label{background}
Given the wide range of available compression techniques, we select four commonly used methods as representative examples to explore their interactions and determine the optimal sequence for applying multiple compression strategies. To systematically investigate the order of compression for CNNs in a general-purpose context, we propose a structured roadmap.

\textbf{The RoadMap to Build Optimal Sequence:}
\begin{itemize}
    \item We start with the exploration of the interactions and the practical application sequence between any two compressions (in Section \ref{sec:interaction}).
    
    \item We then demonstrate that inserting additional compression between any two compressions will not break the application sequence of these two compressions (in Section \ref{sec:additional}).

    \item Built on the application sequence between any two approaches, topological sorting can be used to generate the optimal sequence for applying multiple compressions (in Section \ref{sec:sequence}).
    
    \item After the optimal sequence of multiple compressions is established, the necessity of repeatedly applying any compression is evaluated (in Section \ref{sec:repeat}).
\end{itemize}




As the pioneers in studying the order of compression, this work focuses on ensuring the broadest applicability of the obtained optimal compressed sequence. Therefore, we choose to use the following popular versions of the four compression methods, avoiding advanced variants that are tailored to specific or corner scenarios.

\textbf{Knowledge Distillation}
Neural network distillation, or knowledge distillation, transfers knowledge from a larger, accurate model (teacher) to a smaller, efficient model (student). The student model is trained on both task-specific data and distilled knowledge from the teacher. By mimicking the teacher's output, the student model becomes a compressed version. The implementation in this work follows \cite{tian2019crd}.

\textbf{Early Exit}
Neural network early exit introduces intermediate exit points in the network during inference. If a certain confidence threshold is met, the model exits early and provides the prediction, reducing computation costs and latency. This technique is valuable for real-time applications and resource-limited scenarios, balancing efficiency and accuracy. The implementation in this work follows \cite{passalis2020efficient,li2023predictive}.

\textbf{Pruning}
Neural network pruning reduces the size and complexity of a network by removing less crucial connections, weights, or neurons. This results in a compact network with maintained or improved performance. Pruning can occur during or after training and includes weight, neuron, and filter pruning in CNNs. Benefits include smaller model sizes, reduced memory needs, and faster inference, making pruned models ideal for resource-constrained and real-time applications. This work focuses on channel pruning for hardware optimization and universality \cite{fang2023depgraph}.

\textbf{Quantization}
Neural network quantization reduces the precision of weights and activations, often using 8-bit integers, to decrease memory use and speed up inference. This is ideal for resource-constrained devices. Post-training quantization balances size and efficiency without losing accuracy. This paper uses fixed-point uniform QAT for better accuracy and hardware compatibility \cite{zhou2016dorefa}.
\section{Interaction Between Two Approaches}
\label{sec:interaction}
In this section, we use ResNet34 \cite{resnet} (and CIFAR10 \cite{cifar10} dataset) as a representative convolutional neural network to explore the interactions between two compression approaches. The key insights are also validated on VGG \cite{vgg} and MobileNetV2 \cite{mobilenetv2}. 

\subsection{Complement and Sequence}

In the following subsections, we present the performance of compression (noted as compression ratio) and inference accuracy of the neural network under any two compression approaches. In each scenario, we systematically investigate each compression approach by fine-tuning hyperparameters, which balance the tradeoff between compression ratio and inference accuracy. The performance resulting from the combination of two compression approaches is graphically represented by scatter points, each distinguished by tunable hyperparameters for the respective approach. To maintain methodological consistency throughout the study, we adhere to the following rules:

\begin{itemize}
    \item The BitOps Compression Ratio (BitOpsCR) is used as the metric of compression performance, as we focus more on computation cost in this work and BitOpsCR is suitable for both static and dynamic compression methods. For storage metric, we use Compression Ratio (CR). To standardize the different bit widths between floating point and integer operations, we follow the same BitOps count strategy as \cite{li2019additive} and \cite{liu2021post}. 
    
    \item After each compression operation we will instantly do fine-tuning. To be fair, we keep the same 200 training epochs for both model training (original model, early exit layer, and distillation) and fine-tuning (quantization-aware training and training after pruning) after each compression, while fine-tuning will have 1/10 of the initial learning rate. 
    
    \item For each configuration, we train around 20 cases with the same training hyperparameters but different compression hyper-parameters for each compression method combination pipeline. Each case without early exit will provide one inference accuracy and BitOpsCR sample. Each case with Early Exit will provide several inference accuracy and BitOpsCR samples with different early exit threshold hyperparameters. We believe that doing so maximizes coverage of all cases where compression methods are combined with pipeline. 
\end{itemize}

\begin{figure*}[!htbp]
\begin{center}
\setlength{\abovecaptionskip}{-0.1cm}
\subfigure[DP vs. PD]{
    \label{fig:DP}
    \includegraphics[trim=0cm 0cm 0cm 0cm, clip,width=0.31\textwidth]{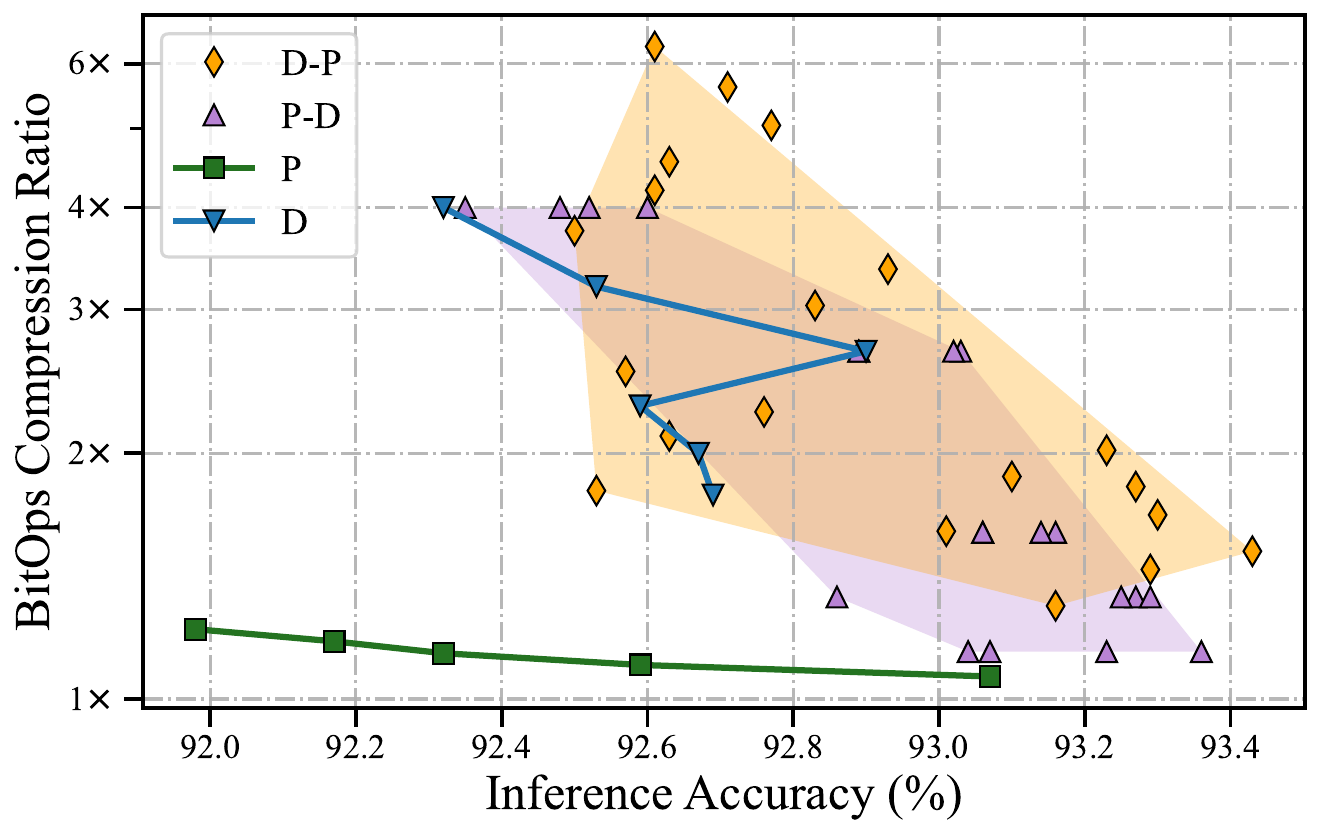}
}
\hspace{-3mm}
\subfigure[DQ vs. QD]{
    \label{fig:DQ} 
    \includegraphics[trim=0cm 0cm 0cm 0cm, clip,width=0.31\textwidth]{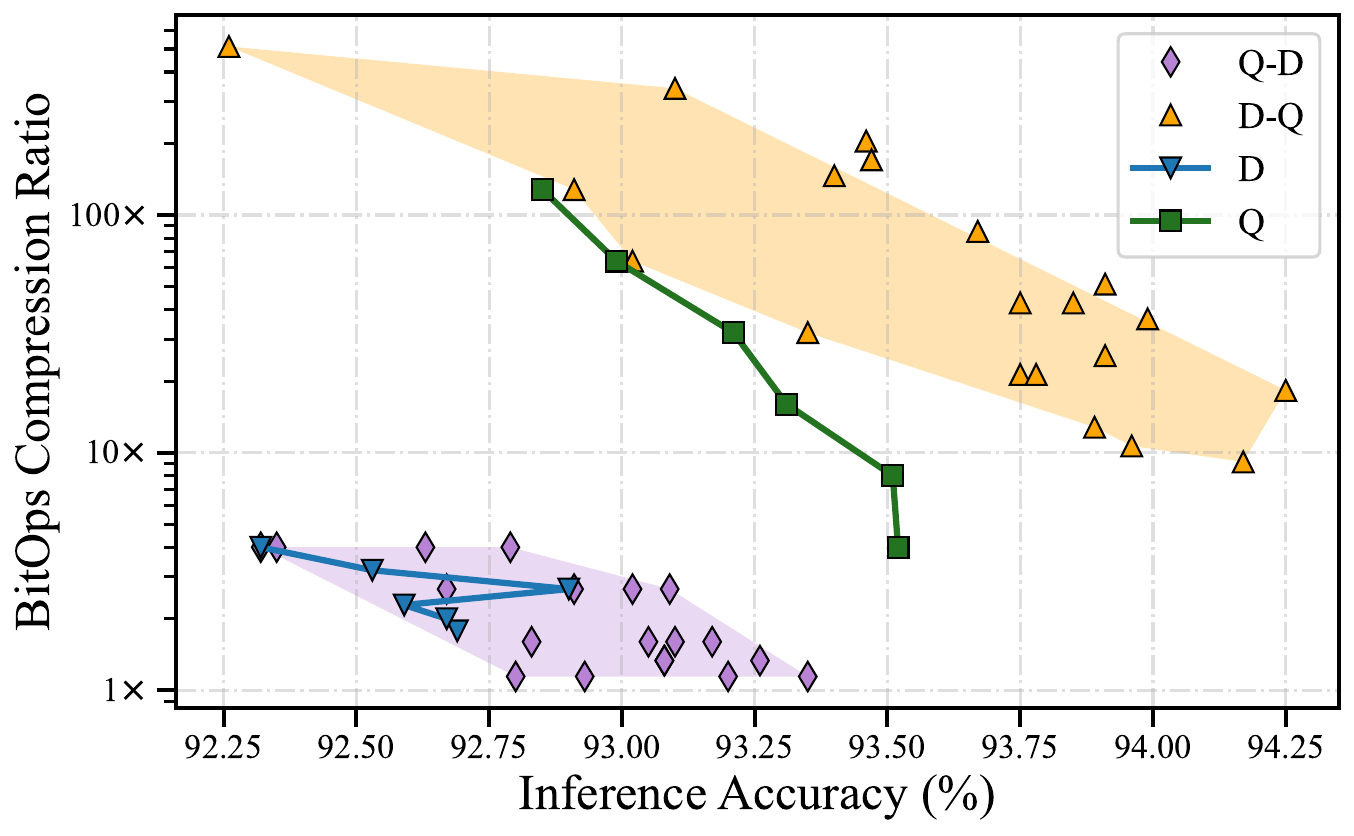}
}
\hspace{-3mm}
\subfigure[DE vs. ED]{
    \label{fig:DE}
    \includegraphics[trim=0cm 0cm 0cm 0cm, clip,width=0.31\textwidth]{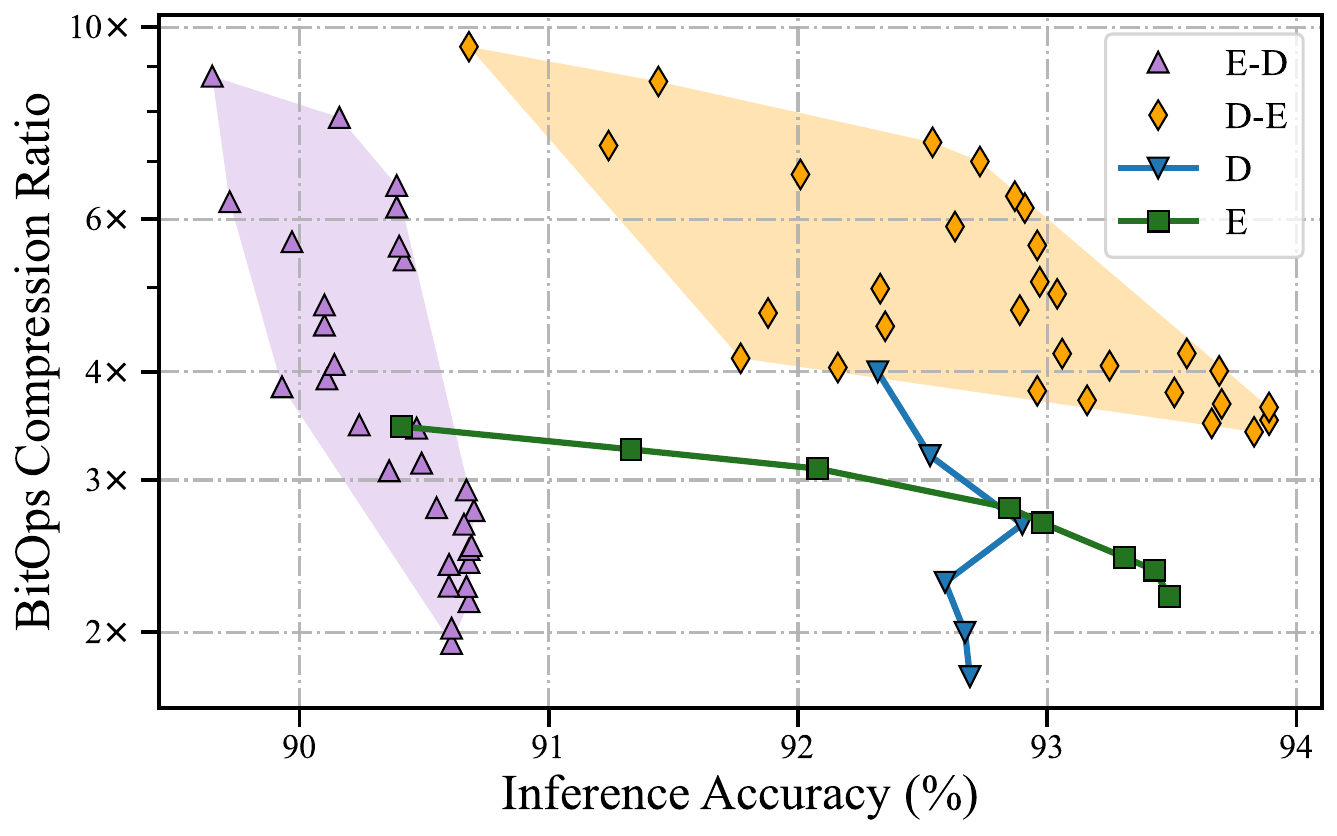}
}
\subfigure[PQ vs. QP]{
    \label{fig:PQ}
    \includegraphics[trim=0cm 0cm 0cm 0cm, clip,width=0.31\textwidth]{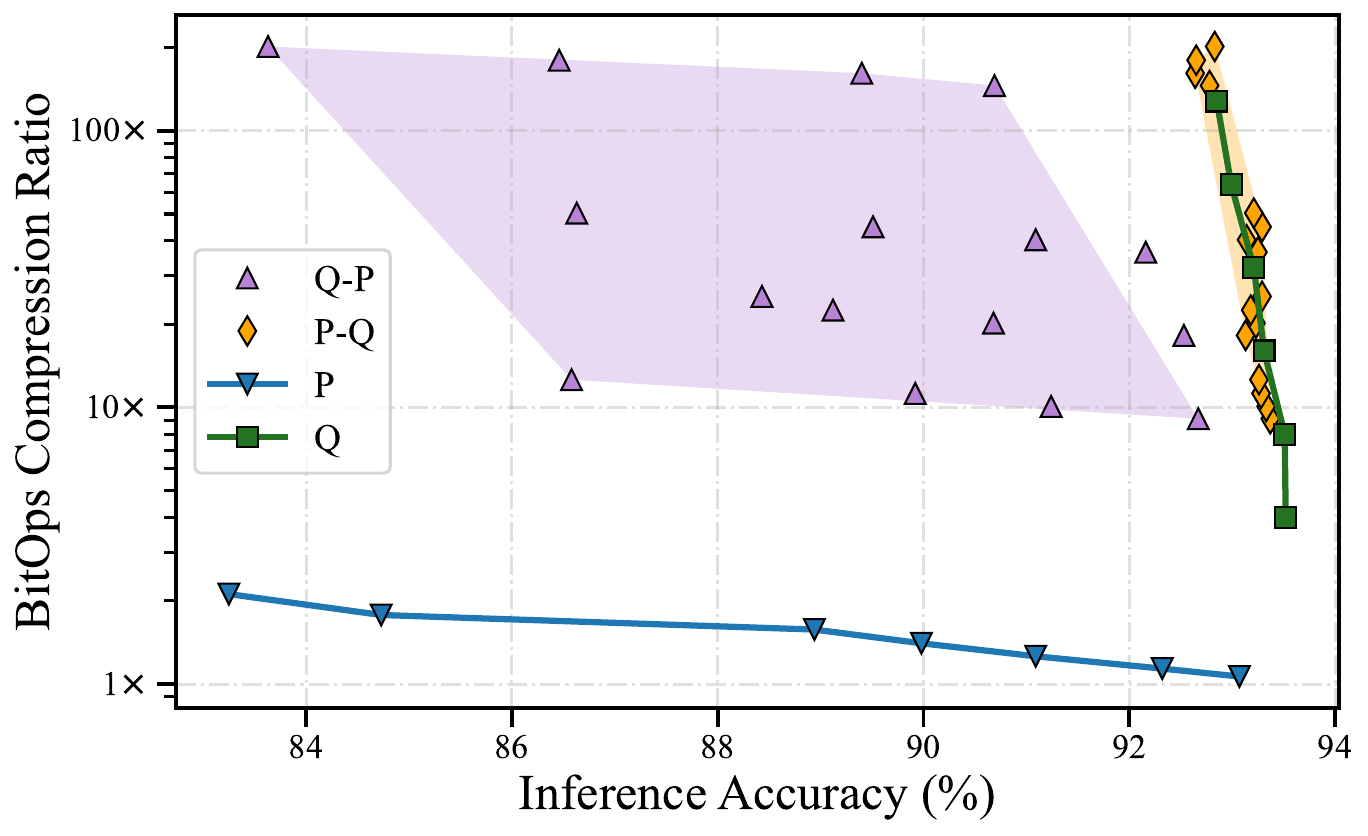}
}
\hspace{-3mm}
\subfigure[PE vs. EP]{
    \label{fig:PE}
    \includegraphics[trim=0cm 0cm 0cm 0cm, clip,width=0.31\textwidth]{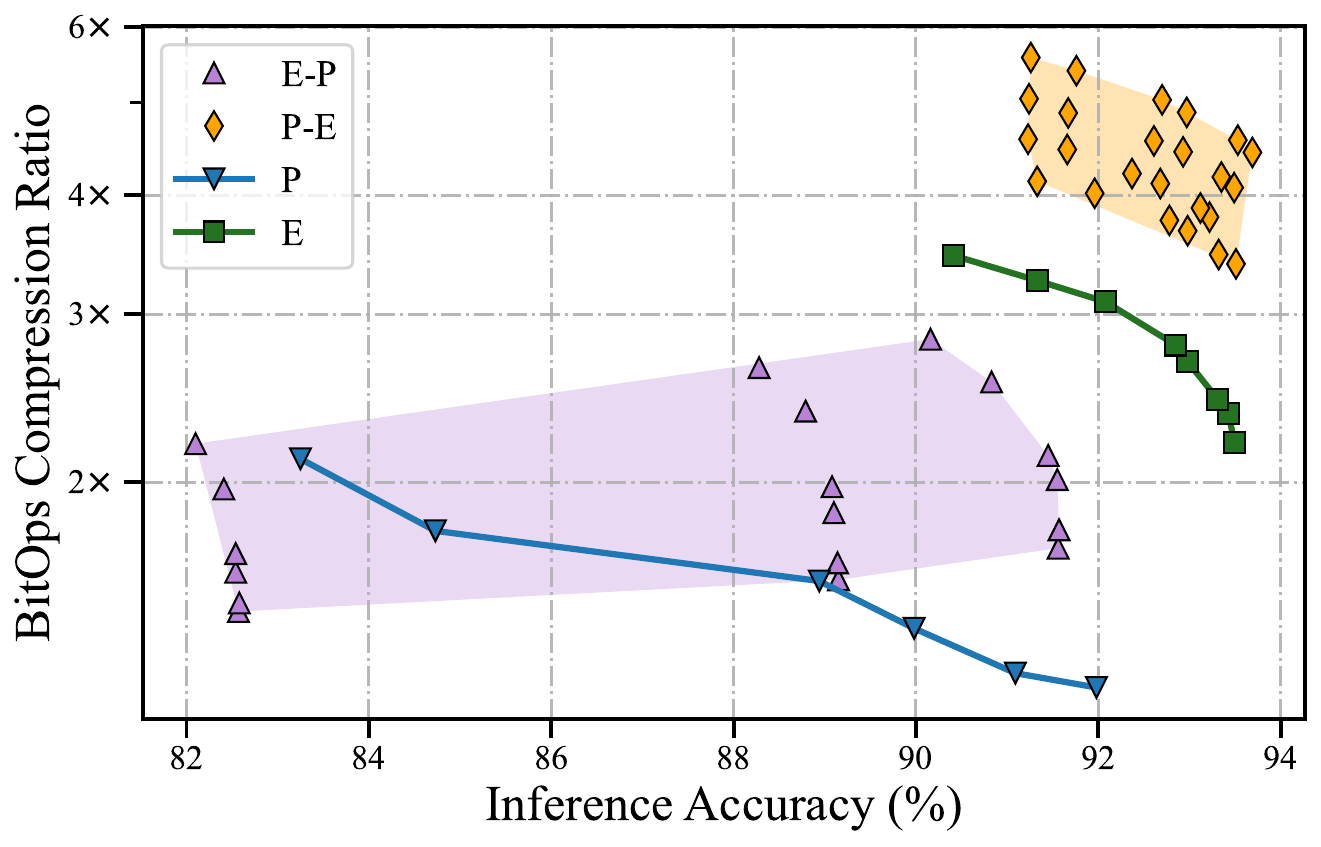}
}
\hspace{-3mm}
\subfigure[QE vs. EQ]{
    \label{fig:QE}
    \includegraphics[trim=0cm 0cm 0cm 0cm, clip,width=0.31\textwidth]{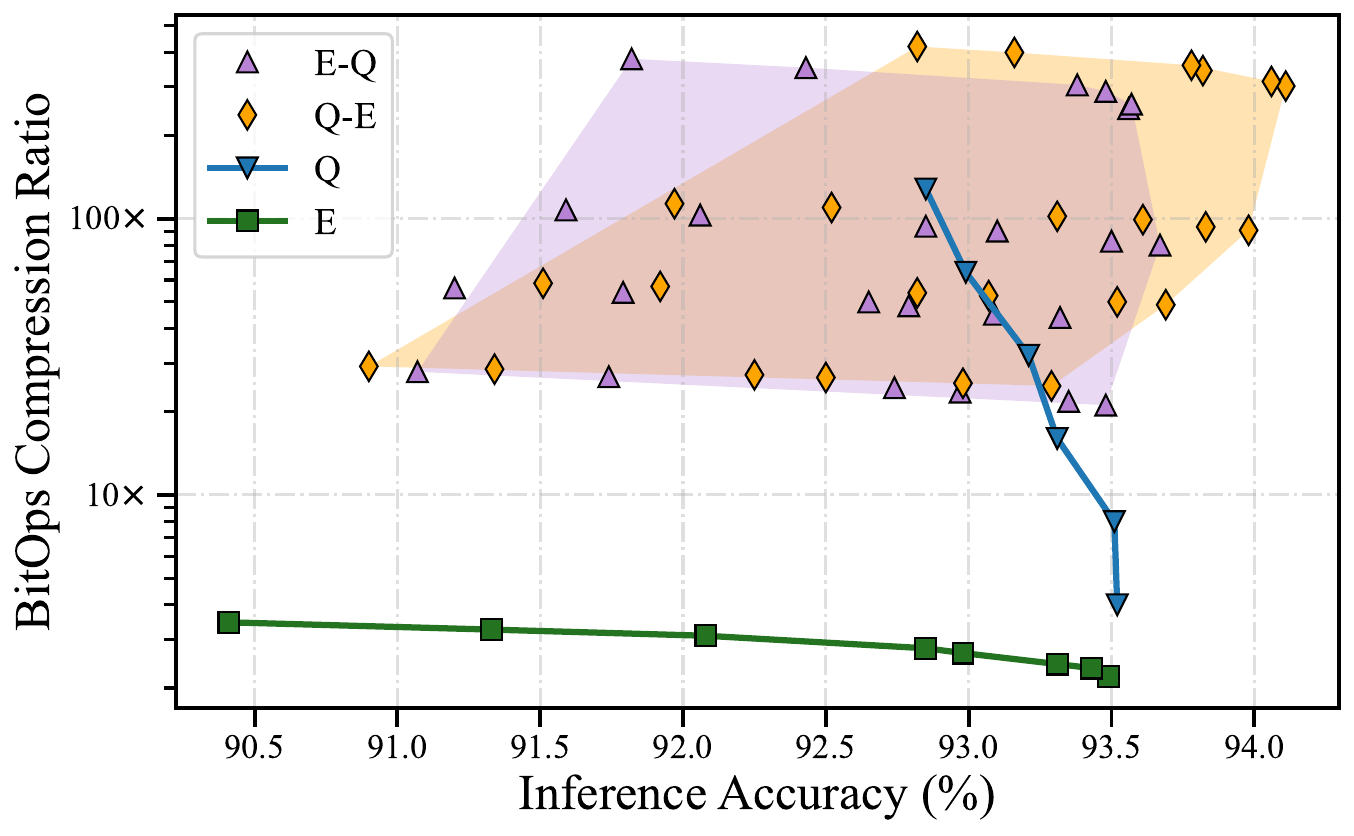}
}
\caption{Significant difference over BitOpsCR-Accuracy trade-off with different compression order.}
\vspace{-4mm}
\label{fig:interaction}
\end{center}
\end{figure*}
\subsubsection{Knowledge Distillation and Pruning}
For the knowledge distillation and pruning shown in Fig. \ref{fig:DP} (DP: After distillation, prune the student model. PD: Before distillation, prune the teacher model first.), the knowledge distillation achieves better BitOps compression ratios given the same inference accuracy. This is due to the fact that uniform channel pruning may not be equally effective for all scenes, as it comes with the price of high hardware compatibility. 

However, when both knowledge distillation and pruning are applied, the sequence of knowledge distillation followed by pruning consistently yields a better BitOps compression ratio while maintaining a comparable inference accuracy. Notably, the scatter plots for first knowledge distillation and then pruning reside mostly in the top-right corner, distinguishing them from the plots corresponding to the reverse sequence. 

Intuitively, any form of compression before distillation may not appear logical. However, in contrast, \cite{f.PD} adopts the order of PD and claims that applying P first helps eliminate redundancies in the student model. Our experiment has shown that their assumption is partially correct, because DP performs even better without any redesigns.



\subsubsection{Knowledge Distillation and Quantization}
For the knowledge distillation and quantization shown in Fig. \ref{fig:DQ} (QD: Before distillation, quantize the teacher model. DQ: Quantize the student model after distillation.), Quantization achieves better BitOps compression ratios given the same inference accuracy. 

When both knowledge distillation and quantization are combined, the sequence of knowledge distillation followed by quantization consistently yields a significantly better BitOps compression ratio while maintaining comparable inference accuracy. Similarly to DP, directly quantizing the student model removes redundancies more efficiently. 

\subsubsection{Knowledge Distillation and Early Exit}
The performance of knowledge distillation and early exit is shown in Fig. \ref{fig:DE} (ED: Train the teacher model's exit layers first, then each teacher model's exit layers together with final Softmax output serve as distillation head of student's exit layers and body layers correspondingly. DE: After distillation, train exit layers on the student model.) 

Knowledge distillation and early exit have relatively close BitOps compression ratios indicated by the cross of these two curves. 
Clearly, first the static compression approach knowledge distillation and then the dynamic compression early exit will achieve much better performance compared to the reverse sequence.
The utilization of the exit layer in exit-aware distillation seems a false statement because the accuracy of the teacher model's last layer return is typically better than most of exit layer returns, so the exit layer return becomes a worse teacher.

 Failures of the train exit layer from the information of the teacher (ED in the figure) show that the information of the student's own body layer is more important for its exit layer. 
 


\subsubsection{Pruning and Quantization}
The compression performance of pruning and quantization is presented in Fig. \ref{fig:PQ} (PQ: Prune + fine-tune first, then quantize + fine-tune the model. QP: Quantize + fine-tune first, then prune and do QAT(Quantization-Aware-Training) as fine-tuning.). 

Benefiting from BitOps as computation cost metric, the quantization achieves better compression than pruning (this is difficult to achieve in hardware, depending on the hardware support for bit operations). 

Quantization limits the amount of information of all neurons in the network, which damages the high neuron resolution requirements of pruning (to select the least important batch of neurons/filters/channels). This is corroborated by our experiment: When applying both pruning and quantization, first the neuron-level compression pruning and then the subneuron-level compression quantization achieve significantly better performance than the first quantization and then pruning. 

\subsubsection{Pruning and Early Exit}
The compression performance of pruning and early exit is presented in Fig.\ref{fig:PE} (EP: Train exit layers first, then prune and fine-tune whole networks including exit layers. PE: Prune and fine-tune the original network first, then train unpruned exit layers.). 

The exit layer allows inference to exit in the middle; therefore, to maintain good exit accuracy, the information density of the exit layer is typically larger than the body layer. It is safer to prune the body layer first as it has more redundancy. The experimental results also confirm this point: When both pruning and early exit are deployed, first pruning (static compression) and then early exit (dynamic compression) will perform significantly better than first early exit than pruning, even though the early exit compresses the neural network from a larger granularity. 



\subsubsection{Quantization and Early Exit}
The performance of the last two compression approaches, quantization and early exit, is shown in Fig. \ref{fig:QE} (EQ: Train exit layers first, then quantize and fine-tune whole networks include exit layers. QE: Quantize and fine-tune original network first, then do QAT from scratch on exit layers.). 

Individually, quantization performs better than early exit. When applying two compression approaches, first the static sub-neuron-level quantization and then the dynamic architectural level early exit will achieve better compression performance, especially with the requirement of high neural network inference accuracy.

Unlike pruning, the exit layer itself should accept a quantized activation value. As a result, in QE even Q first, the later created exit layer should be quantized from the beginning and do QAT form scratch. The advantage of DE, QE, and PE indicates that fine-tuning on early exit layers is not as effective as on original network for accuracy recovery.


\subsection{Summary of Interactions}
Based on the explorations of any two compression techniques, we summarize the key insights here.
\begin{itemize}
    \item A clear complementary feature is observed. Applying two compressions with the optimal sequence can achieve better compression performance compared to an individual single compression.
    \item The sequence of applying two compression approaches will directly impact the compression rate and inference accuracy.
    \item There is no evidence to support that first applying the compression with a large compression ratio and then applying the compression with a smaller compression ratio is an optimal sequence.
    \item First applying the compression working on large granularity then the compression on small granularity, or first applying the static compression then the dynamic compression can achieve better performance. 
\end{itemize}

\section{Adding Additional Compression}
\label{sec:additional}
In this section, we aim to validate the integration of an additional compression step without disrupting the established interaction and sequence delineated in Section \ref{sec:interaction}. The insertion of compression, whether positioned before or after the previously outlined two approaches, inherently modifies the input or output models. However, the impact on established flow is comparatively less disruptive than the introduction of an additional compression step between the two already established approaches. 
\begin{figure*}[!htbp]
\begin{center}
\setlength{\abovecaptionskip}{-0.1cm}
\subfigure[PQE vs. EQP]{%
    \includegraphics[trim=0cm 0cm 0cm 0cm, clip,width=0.32\textwidth]{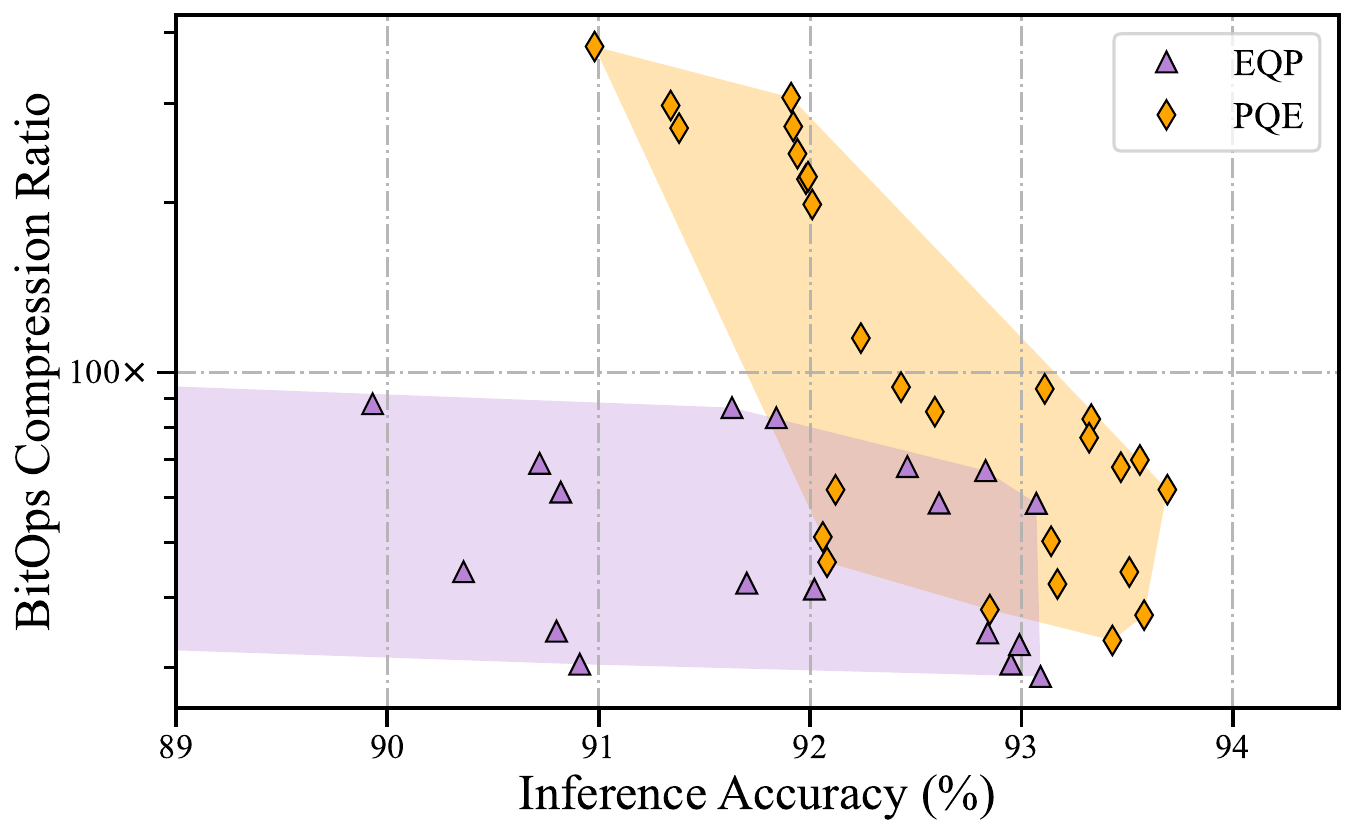}
}
\hspace{-3mm}
\subfigure[PEQ vs. QEP]{
    \includegraphics[trim=0cm 0cm 0cm 0cm, clip,width=0.32\textwidth]{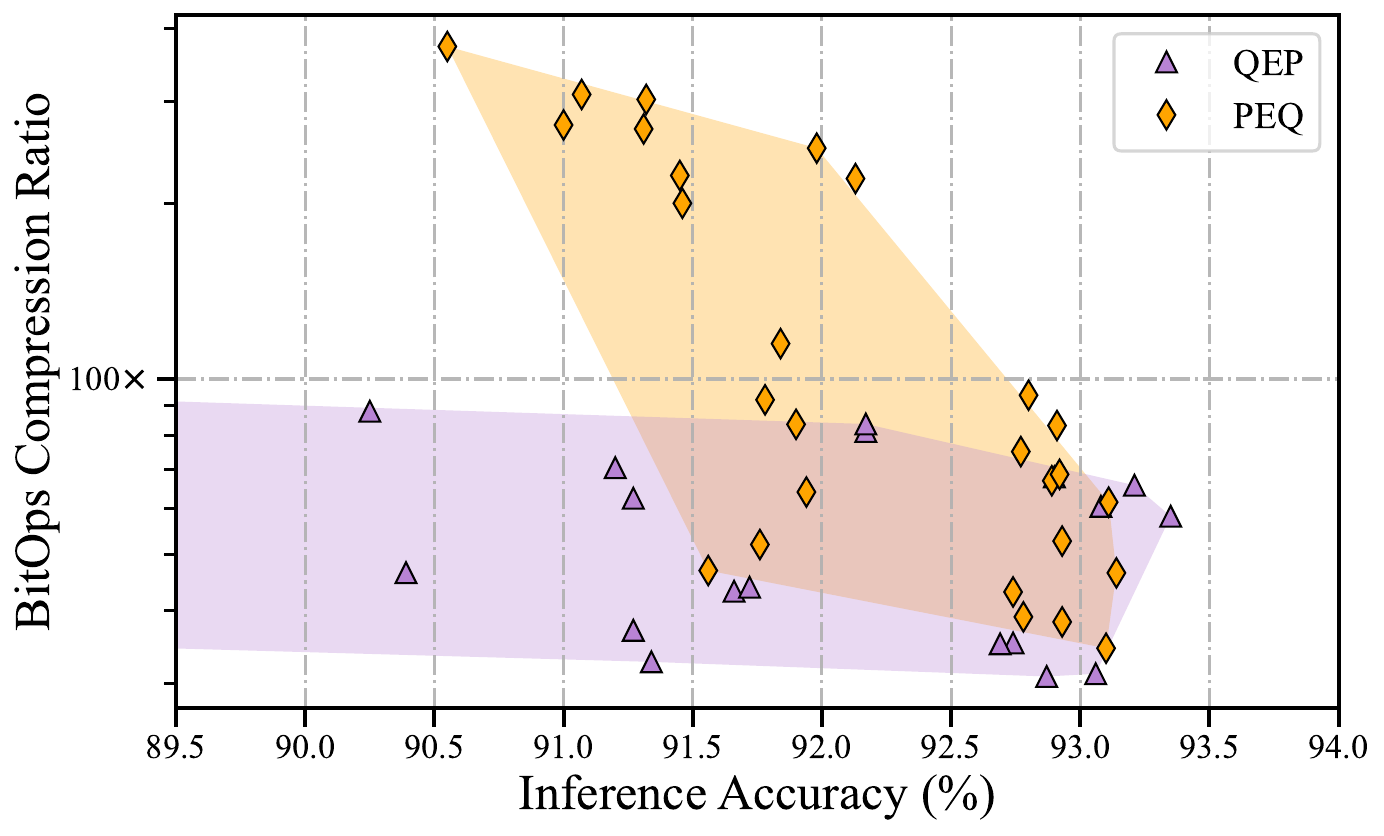}
}
\hspace{-3mm}
\subfigure[QPE vs. EPQ]{
    \includegraphics[trim=0cm 0cm 0cm 0cm, clip,width=0.32\textwidth]{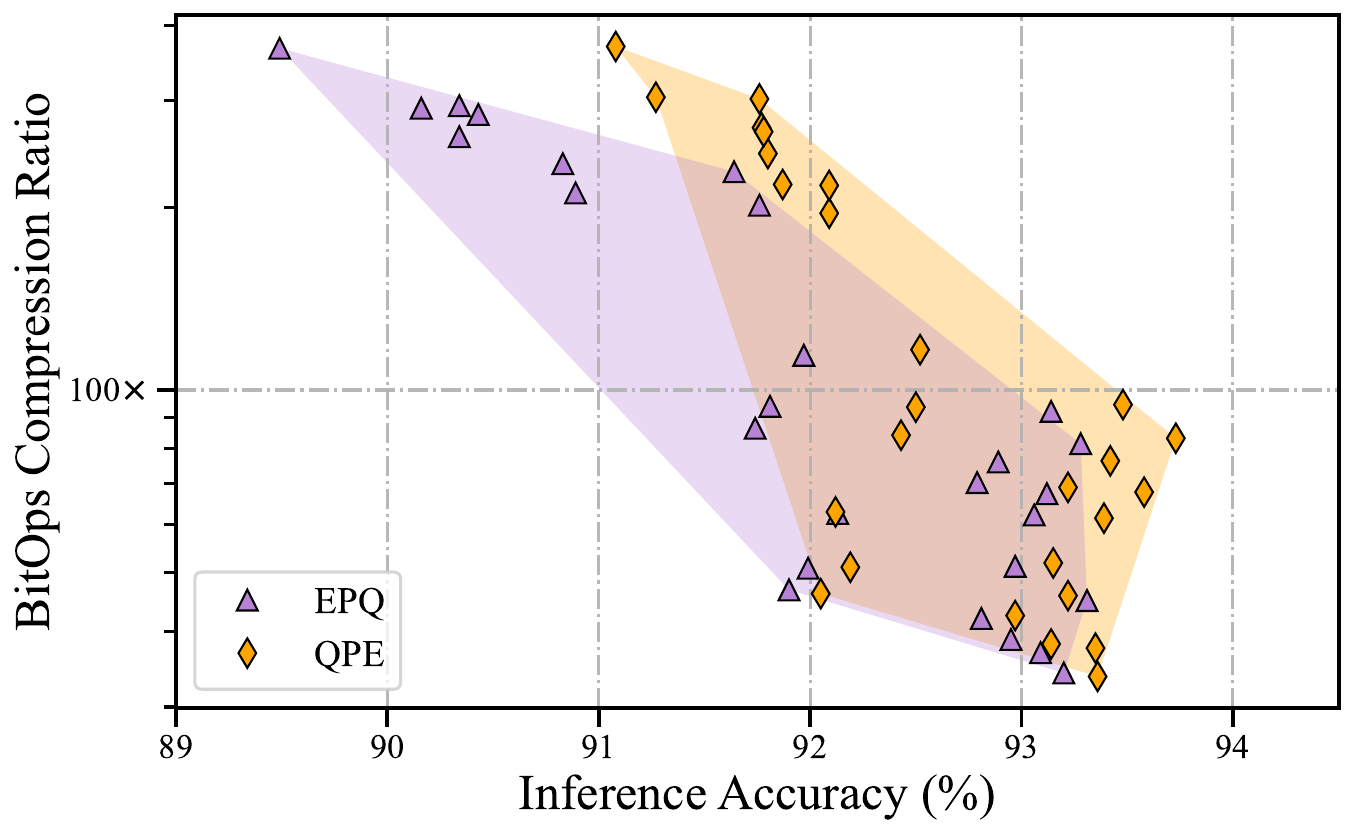}
}
\caption{The established sequence remains unaffected by inserting one more compression approach in the middle.}
\label{fig:addition}
\end{center}
\vspace{-3mm}
\end{figure*}
Given the precedence of knowledge distillation (over other approaches) established by the previous section's discovery, we investigate the integration of Quantization/Pruning/Early Exit into the other two compression techniques. Fig. \ref{fig:addition} illustrates the performance when inserting Quantization/Pruning/Early Exit between the other two compressions. In particular, regardless of whether we introduce quantization, pruning, or early exit as an additional compression step, the established sequence (P[Q]E, P[E]Q, Q[P]E) remains compatible with that presented in the previous section (PE, PQ, QE).

\section{Combinational Sequence Law}
\label{sec:sequence}
Having established the sequence between each pair of compression approaches and validated that adding additional compression approaches will not break the sequence, we finally build the combinational sequence for applying multiple compressions using topological sorting. As prerequisites for topological sorting, it is imperative to ensure the absence of cycles within the sorting process. Accordingly, we make the assumption that each compression approach is used only once in this section. In the following section, we will provide a brief discussion of the iterative application of compression approaches.

Following the established sequence in Section \ref{sec:interaction} and the topological sorting, the optimal combination sequence to apply the compression approaches is 

\begin{tiny}
\begin{equation*}
    \begin{matrix} \textbf{\small{Distillation}}\\{(Static)}\\ {(Architecture)}\end{matrix}\rightarrow\begin{matrix} \textbf{\small{Pruning}}\\{ (Static)}\\ { (Neuron)}\end{matrix}\rightarrow\begin{matrix} \textbf{\small{Quantization}}\\{ (Static)}\\ { (Sub-Neuron)}\end{matrix}\rightarrow\begin{matrix} \textbf{\small{Early Exit}}\\{ (Dynamic)}\\ { (Architecture)}\end{matrix}
\end{equation*}
\end{tiny}

\textit{The topological sorting of the compression approach reveals a consistent adherence to the principles of transitioning from static to dynamic and progressing from a large granularity perspective to a small granularity one.}

\begin{figure}[t]
\centering
\includegraphics[width=0.33\textwidth]{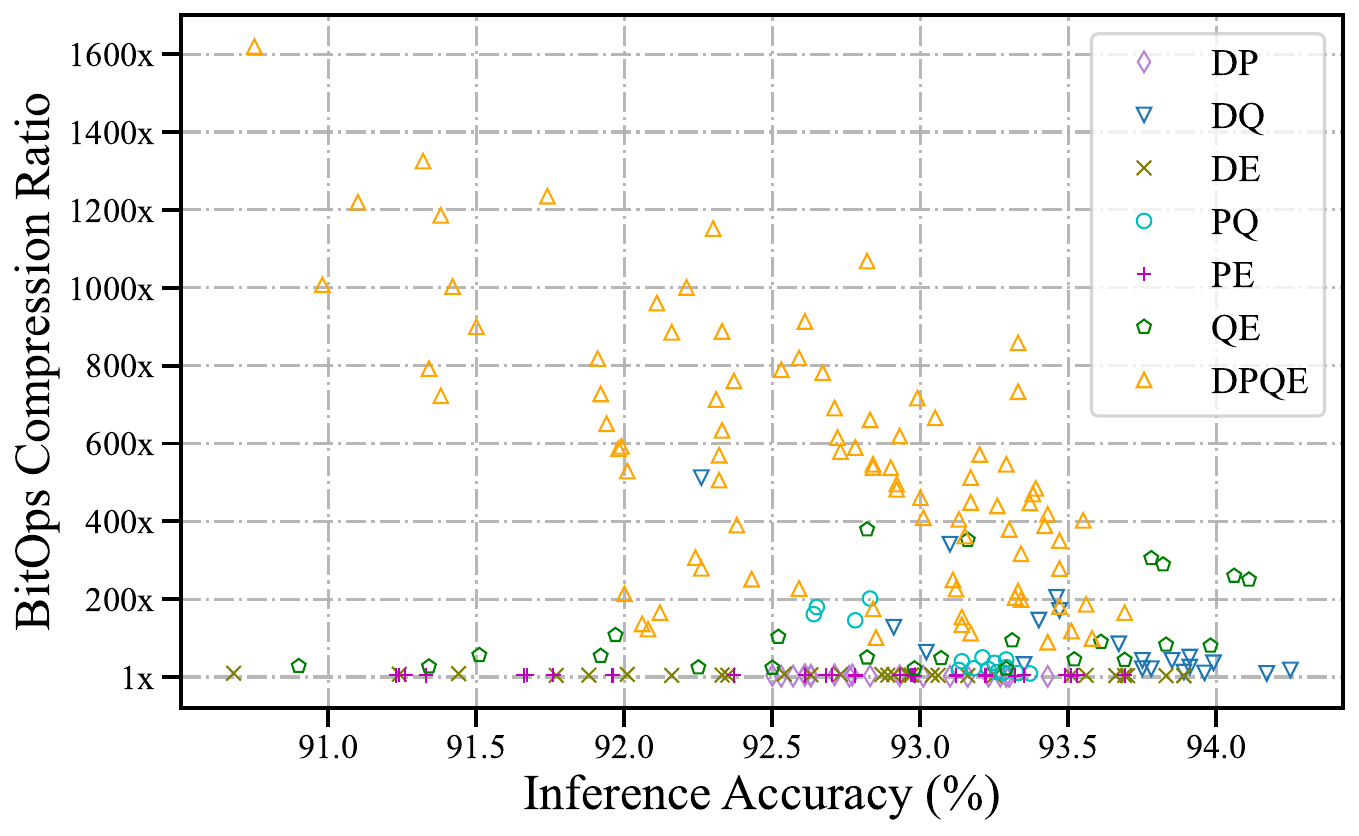}
\caption{Combinational compression with previously established two approaches.}
\label{fig:DPQE-dual}
\vspace{-3mm}
\end{figure}
Fig. \ref{fig:DPQE-dual} presents the BitOps compression ratio and network accuracy using the full combinational sequence and the previously established two approaches in Sec. \ref{sec:interaction}. Compared with the previously established two approaches, the combinational sequence may have slightly lower (less than 1\%) reachable network inference accuracy. This is because the combination sequence incorporates pruning that limits the inference accuracy. If we compare the BitOps compression ratio within the reachable region of the combinational sequence (i.e., $\leq$ 93.5\%), the combination sequence achieves a significantly improved BitOps compression ratio. For example, it can compress the neural network by about {1000}$\times$ with a network accuracy of {92.8\%} (i.e., 0.6\% loss over the original ResNet34\cite{he2015deep} which has an accuracy of 93.42\%).

\begin{table}[t]
\setlength{\tabcolsep}{5pt}
\renewcommand{\arraystretch}{0.8}
  \centering
\begin{scriptsize}
  \caption{BitOps compression ratio of all D-started compression sequences (The original ResNet34's accuracy is 93.42\%).}
    \begin{tabular}{ccccccc}
    \toprule
    \makecell{Acc. Loss} & \textbf{DPQE} & DQPE    & DPEQ    & DQEP    & DEPQ    & DEQP \\
    \midrule
    $\leq0.2\%$ & \boldmath $858\times$ & $555\times$ & -       & $155\times$ & $217\times$ & - \\
    $\leq0.6\%$ & \boldmath $1068\times$ & $667\times$ & $221\times$ & $181\times$ & $245\times$ & $178\times$ \\
    $\leq1.0\%$ & \boldmath $1068\times$ & $923\times$ & $250\times$ & $181\times$ & $245\times$ & $181\times$ \\
    $\leq2.0\%$ & \boldmath $1235\times$ & $1238\times$ & $664\times$ & $223\times$ & $611\times$ & $231\times$ \\
    \bottomrule
    \end{tabular}%
  \label{tab:compare_four}%
\end{scriptsize}
\vspace{-3mm}
\end{table}%

Table \ref{tab:compare_four} provides a comprehensive comparison of the maximum achievable BitOps compression ratios under the optimal sequence law and other combination sequences, considering various tolerable accuracy losses. Given the demonstrated effectiveness of distillation as the initial compression step, we specifically evaluate sequences that start with distillation. The optimal sequence law, denoted as DPQE, consistently attains significant BitOps compression ratios across all acceptable accuracy loss thresholds. Comparatively, a sequence with closer alignment to the optimal sequence law (i.e., DQPE) achieves a secondary BitOps compression ratio. However, sequences such as DQEP and DEQP, which deviate significantly from the optimal sequence law, exhibit notably lower BitOps compression ratios. The performance across different sequences also proves the importance of the optimal sequence law in applying multiple compressions. 
\vspace{-1mm}

\section{Repeating the Compression}
\label{sec:repeat}

\begin{figure}[t]
\centering
\includegraphics[width=0.38\textwidth]{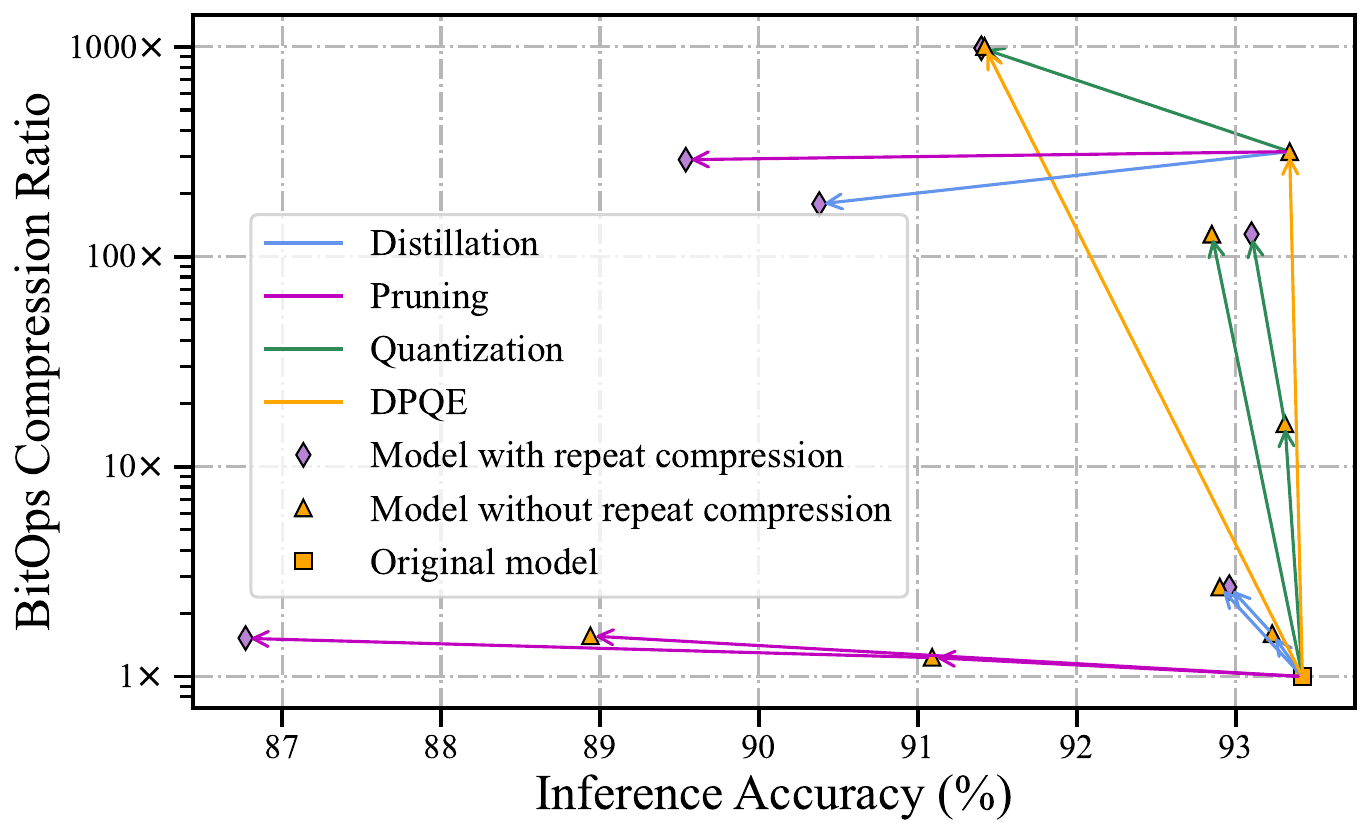}
\caption{Repeating compressions.}
\label{fig:repeat}
\vspace{-3mm}
\end{figure}

As a necessary step in exploring the optimal combination sequence, we investigated the repeated application of compression techniques. Our exploration encompasses two scenarios: one involves the continuous repetition of a single compression method, while the other repeats a specific compression technique after employing the optimal sequence formed by combining all approaches. As the early exit is a dynamic compression at runtime that cannot be repeated, we do not consider the repeating of early exit. Fig. \ref{fig:repeat} presents the compression performance of repeatedly using one compression approach. In the figure, each marker represents a network model. The triangle marker means the model is not repeatedly compressed while the rhombus means the model is repeatedly compressed by one method. The line represents the application of compression, and the color of the line indicates the method of compression. 

To assess the impact of continuous repetition of a single compression method, we additionally depict the compression performance using the corresponding single compression method with more aggressive hyperparameters to signify a higher BitOps compression ratio with increased accuracy loss. Notably, the results show that repeating distillation twice yields a performance similar to applying distillation once with more aggressive hyperparameters. Conversely, repeating pruning twice leads to a diminished performance compared to applying pruning once with more aggressive hyperparameters. Repeating quantization twice results in marginally improved performance compared to applying quantization once with more aggressive hyperparameters.

Starting from the combinational applied four compression approach with the optimal sequence (noted as DPQE), repeating either distillation or pruning will not decrease the compression performance. Repeating the quantization will increase the BitOps compression ratio but an obvious accuracy loss is reported. Although continuous repetition of quantization could lead to better compression, repeating the quantization after four compression approaches with the optimal sequence breaks the sequence of two approaches in the previous sections where the quantization (static compression) should be applied before the early exit (dynamic compression). Therefore, repeating the quantization after the four compression approaches with the optimal sequence will not lead to an improved compression performance. 

Therefore, we can conclude that repeating the compression process does not significantly enhance compression performance, except in the case of continuous repetition of quantization. The importance of maintaining an optimal sequence becomes evident. After employing the four compression approaches in the optimal sequence, repeating a single compression method may disrupt this sequence and result in a reduction of compression performance.

\section{Evaluation}
\label{sec:evaluation}
\begin{figure*}[!htbp]
\begin{center}
\setlength{\abovecaptionskip}{-0.1cm}
\subfigure[VGG19]{
    \label{fig:vggchange}
    \includegraphics[trim=0cm 0cm 0cm 0cm, clip,width=0.3\textwidth]{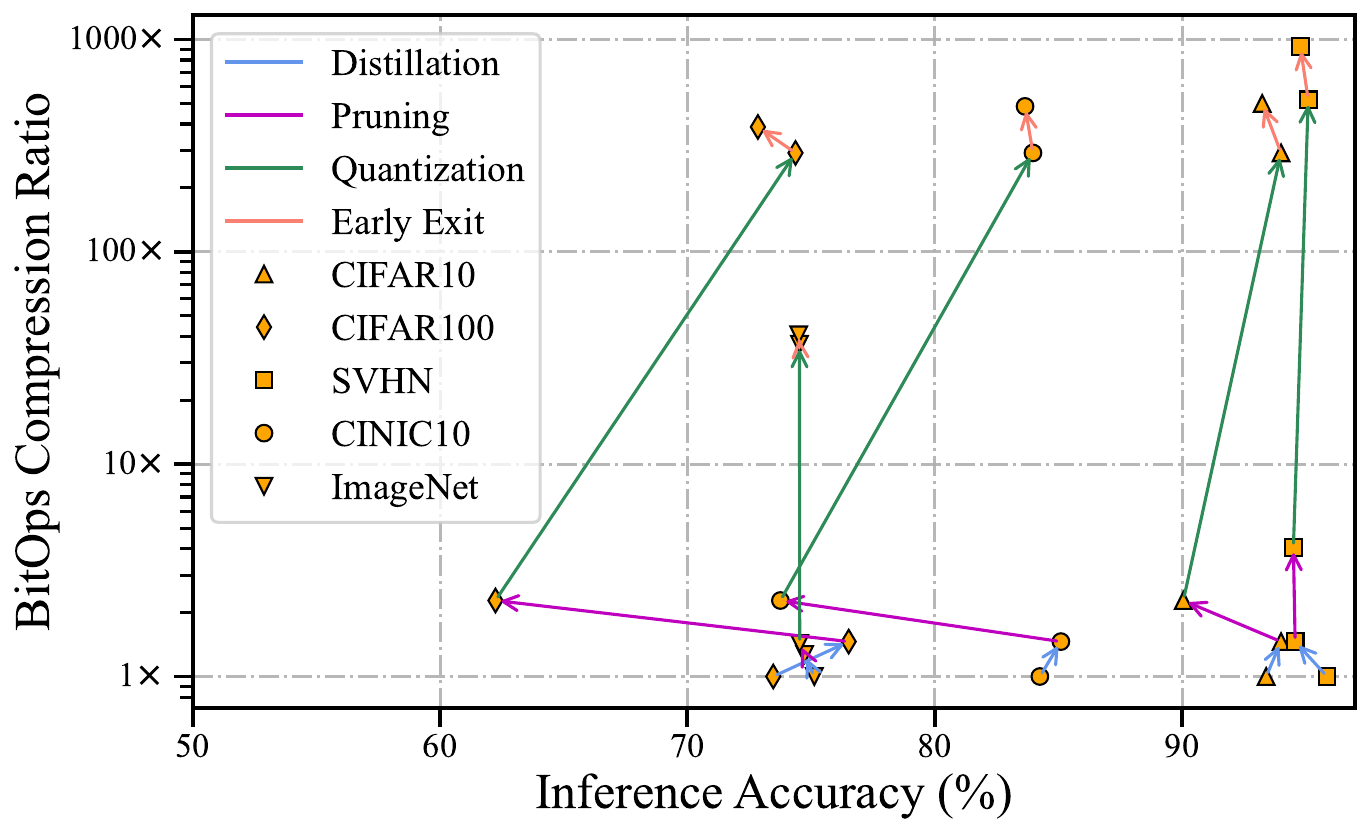}
}
\hspace{-3mm}
\subfigure[ResNet34]{
    \label{fig:resnetchange}
    \includegraphics[trim=0cm 0cm 0cm 0cm, clip,width=0.3\textwidth]{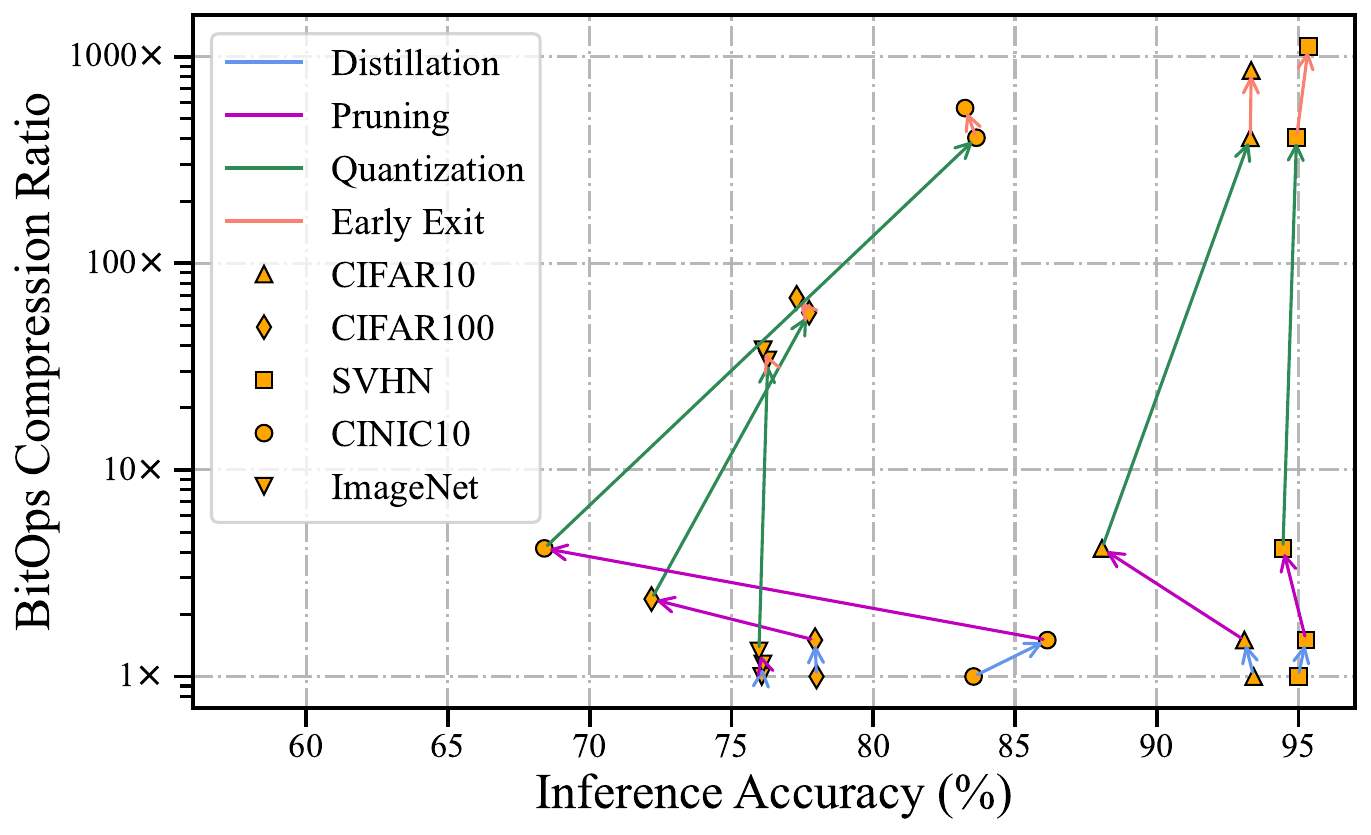}
}
\hspace{-4mm}
\subfigure[MobileNetV2]{
    \label{fig:mobilenetchange}
    \includegraphics[trim=0cm 0cm 0cm 0cm, clip,width=0.3\textwidth]{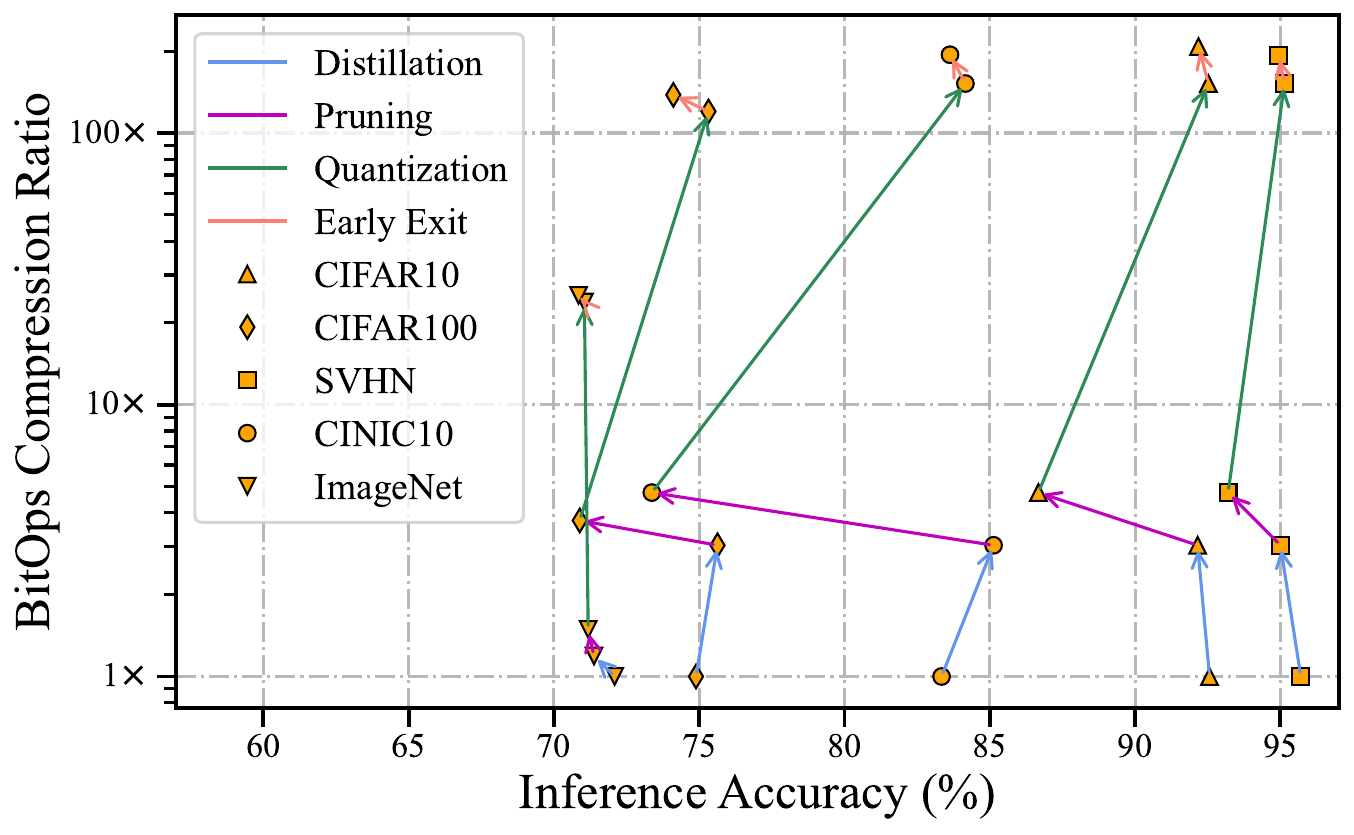}
}
\caption{Performance change with applying each compression.}
\label{fig:change}
\end{center}
\vspace{-3mm}
\end{figure*}
We perform an extensive evaluation of end-to-end performance on popular CNN network architectures, namely VGG-19 \cite{vgg}, ResNet34 \cite{resnet} and MobileNetV2 \cite{mobilenetv2}, across diverse commonly used benchmark datasets, including CIFAR-10, CIFAR-100 \cite{cifar10}, SVHN \cite{svhn}, CINIC10 \cite{darlow2018cinic} and ImageNet-1k(ILSVRC-2012) \cite{ILSVRC15}. For Cifar-style datasets, to ensure compatibility, we used a modified version of MobileNetV2 introduced by \cite{RMNv2}. Given that MobileNetV2 scales primarily by width (as opposed to the depth of ResNet), the student model of MobileNetV2 maintained the same depth while featuring a reduced width (i.e., a lower channel number on each layer). 

Tables \ref{tab:totalCR} summarize the compression performance of the proposed Order of Compression on VGG19, ResNet34, and MobileNetV2 models in various benchmarks. The Order of Compression achieves remarkable compression ratios ranging from hundreds to over 1000 times. Upon closer examination of the Order of Compression's performance on each dataset, a clear pattern emerges, indicating that more substantial compression is attainable for simpler classification tasks such as CIFAR10, SVHN, and CINIC10. Conversely, for the intricate classification task of CIFAR100 and ImageNet, compression ratios tend to be smaller, accompanied by the possibility of an increase in accuracy loss. This observed trend aligns with the compression performance of each individual compression technique presented in the previous sections. 
\begin{table}
\setlength{\tabcolsep}{3.0pt}
\renewcommand{\arraystretch}{0.8}
  \centering
\begin{scriptsize}
  \caption{Accuracy change and CRs on VGG19, ResNet34 and MobileNetV2.}
    \begin{tabular}{cccccc}
    \toprule
    Model & Dataset & \makecell[c]{Original \\Acc.(\%)} & \makecell[c]{Compressed \\Acc.(\%)} & BitOpsCR & CR \\
    \midrule
    \multirow{5}{*}{VGG19} & CIFAR10 & $93.40$   & $93.24(-0.16)$   & $500\times$   & $105\times$ \\
     & CIFAR100 & $73.47$   & $72.85(-0.62)$ &  $387\times$  & $104\times$ \\
     & SVHN   & $95.86$   & $94.78(-1.08)$   & $927\times$ & $185\times$ \\
     & CINIC10 & $84.25$   & $83.65(-0.60)$   & $484\times$ &$105\times$ \\
     & ImageNet & $75.13$   & $74.50(-0.63)$   & $41\times$ & $9\times$ \\
    \midrule
    \multirow{5}{*}{ResNet34} & CIFAR10 & $93.42$   & $93.33(-0.09)$   & $859\times$   & $114\times$ \\
     & CIFAR100 & $78.00$   & $77.30(-0.70)$ &  $68\times$  & $14\times$ \\
     & SVHN   & $94.99$   & $95.35(+0.36)$   & $1121\times$ & $114\times$ \\
     & CINIC10 & $83.54$   & $83.24(-0.30)$   & $564\times$ &$114\times$ \\
     & ImageNet & $76.07$ & $76.12(+0.05)$ & $38\times$ & $8\times$ \\
    \midrule
    \multirow{5}{*}{MobileNetV2} & CIFAR10 & $92.56$   & $92.18(-0.38)$   & $208\times$   & $35\times$ \\
     & CIFAR100 & $74.88$   & $74.11(-0.78)$ &  $138\times$  & $23\times$ \\
     & SVHN   & $95.70$   & $94.94(-0.76)$   & $193\times$ & $35\times$ \\
     & CINIC10 & $83.34$   & $83.63(+0.29)$   & $194\times$ &$35\times$ \\
     & ImageNet & $71.63$   & $70.85(-0.78)$ & $25\times$ & $5\times$ \\
    \bottomrule
    \end{tabular}%
  \label{tab:totalCR}%
\end{scriptsize}
\vspace{-3mm}
\end{table}

\begin{table*}[!h]
  \centering
\setlength{\tabcolsep}{2.2pt}
\renewcommand{\arraystretch}{0.8}
\begin{threeparttable}
\begin{scriptsize}
  \caption{Comparing the Order of Compression (noted as Ours. DPQE) with other compression methods involving multiple different compression techniques.}
        \begin{tabular}{cccccccccccc}
    \toprule
    Dataset & Model   & Method\tnote{1}  & Acc. (\%) & BitOpsCR & CR      & Dataset & Model   & Method  & Acc. (\%) & BitOpsCR & CR \\
    \midrule
    \multirow{14}[7]{*}{ImageNet} & \multirow{4}[3]{*}{ResNet34} & d. CEA-MOP & 72.9(-1.11) & 1.70 & 1.49 & \multirow{2}[1]{*}{CIFAR100} & \multirow{2}[1]{*}{VGG19} & c. Predictive E+Q & 62(-2)  & 20.94 & - \\
            & & j. Quantized Distill. & 73.10(-0.21) & - & 3.97 & & & \textbf{Ours. DPQE} & \textcolor[rgb]{0, .69, .314}{72.85(-0.62)} & \textbf{387.26 } & \textbf{104.45 } \\
\addlinespace[1pt]\cline{7-12}\addlinespace[1pt]
            & & o. APQ  & \textcolor[rgb]{0, .69, .314}{75.1(+0.1)}\tnote{2} & 8 & - & \multirow{17}[9]{*}{CIFAR10} & \multirow{2}[1]{*}{ResNet18} & a. OICSR-GL & \textcolor[rgb]{0, .69, .314}{94.26(-0.40)} & 7.41 & 10.98 \\
            & & \textbf{Ours. DPQE} & \textcolor[rgb]{0, .69, .314}{76.12(+0.05)} & \textbf{38.15} & 8.42 & & & b. PCQAT & 85(-3)  & - & 11.82 \\
\addlinespace[1pt]\cdashline{2-6}[1pt/2pt] \cdashline{8-12}[1pt/2pt]\addlinespace[1pt]
            & \multirow{4}[1]{*}{ResNet50} & a. OICSR-GL & \textcolor[rgb]{ 0,  .69,  .314}{75.76(-0.74)} & 32      & -       &         & \multirow{4}[1]{*}{ResNet34} & c. Predictive E+Q & 82(-3)  & 190.48  & - \\
            &         & m. HFPQ & 73.06(-3.09) & 3.65 & \textbf{15}      &         &         & d. CEA-MOP & 92.01(-0.62) & 1.70    & 1.62  \\
            &         & h. HMC  & \textcolor[rgb]{ 0,  .69,  .314}{75.55(-0.77)} & -       & 5.35    &         &         & e. LQP  & 92.43(-0.43) & 30.52   & 22.81  \\
            &         & l. Hybrid Search & \textcolor[rgb]{ 0,  .69,  .314}{75.74(-0.27)} & 2.38 & -       &         &         & \textbf{Ours. DPQE} & \textcolor[rgb]{ 0,  .69,  .314}{93.33(-0.09)} & \textbf{858.70 } & \textbf{113.85 } \\
\addlinespace[1pt]\cline{2-6} \cdashline{8-12}[1pt/2pt]\addlinespace[1pt]
            & \multirow{2}[1]{*}{VGG16} & \multicolumn{1}{l}{n. Deep Compression} & 68.83(-0.33) & -       & \textbf{49}      &         & ResNet110 & \multirow{2}[1]{*}{f. PD} & 93.00(-1.27) & -       & 4.70  \\
            &         & k. Smart-DNN+ & 70.83(+0.03) & -       & 4.8     &         & ResNet164 &         & \textcolor[rgb]{ 0,  .69,  .314}{93.70(-0.82)} & -       & 3.63  \\
\addlinespace[1pt]\cdashline{2-6}[1pt/2pt] \cdashline{8-12}[1pt/2pt]\addlinespace[1pt]
            & VGG19   & \textbf{Ours. DPQE} & \textcolor[rgb]{ 0,  .69,  .314}{74.5(-0.63)} & \textbf{40.64} & 8.62 & & \multicolumn{1}{c}{\multirow{3}[1]{*}{{\makecell[c]{ResNet56\\(Cifar ver.)}}}} & g. KDDP SN & 81.93(-6.16) & - & 20.37 \\
\addlinespace[1pt]\cline{2-6}\addlinespace[1pt]
            & \multicolumn{1}{c}{\multirow{3}[4]{*}{{\makecell[c]{Mobile-\\NetV2}}}} & l. Hybrid Search & \textcolor[rgb]{ 0,  .69,  .314}{71.3(-0.7)} & 1.37 & -       &         &         & h. HMC  & 92.54(-0.89) & -       & 5.35  \\
            &         & o. APQ  & \textcolor[rgb]{ 0,  .69,  .314}{72.1(+0.3)} & \multicolumn{1}{c}{1.45} & -       &         &         & i. DDSL & \textcolor[rgb]{ 0,  .69,  .314}{93.98(+0.17)} & 4.15 & 4.00 \\
\addlinespace[1pt]\cline{8-12}\addlinespace[1pt]
            &         & \textbf{Ours. DPQE} & \textcolor[rgb]{ 0,  .69,  .314}{70.85(-0.78)} & \textbf{25.24} & \textbf{4.62} & & \multirow{4}[3]{*}{VGG16} & e. LQP  & \textcolor[rgb]{0, .69, .314}{93.34(-0.21)} & - & 60.99  \\
\addlinespace[1pt]\cline{1-6}\addlinespace[1pt]
            \multirow{5}[3]{*}{CIFAR100} & \multirow{4}[2]{*}{ResNet34} & c. Predictive E+Q & 60(-2)  & 86.49   & -       &         &         & k. Smart-DNN+ & 92.60(+0.00) & -       & 7.07  \\
            &         & h. HMC  & 67.88(-0.94) & -       & 5.56    &         &         & l. Hybrid Search & \textcolor[rgb]{ 0,  .69,  .314}{93.95(-0.52)} & 4.50    & 9.79  \\
            &         & \textbf{Ours. DPQE}\tnote{3} & \textcolor[rgb]{ 0,  .69,  .314}{77.3(-0.7)} & 68.07   & 14.21   &         &         & m. HFPQ & 92.47(-1.11) & 7.58    & 30.00  \\
\addlinespace[1pt]\cdashline{8-12}[1pt/2pt]\addlinespace[1pt]
            &         & \textbf{Ours. DPQE} & 73.5(-4.5) & \textbf{306.88 } & \textbf{64.17 } &         & \multirow{2}[3]{*}{VGG19} & c. Predictive E+Q & 87(-1)  & 34.71   & - \\
\addlinespace[1pt]\cline{2-6}\addlinespace[1pt]
            & VGG16   & k. Smart-DNN+ & 66.01(-0.03) & -       & 6.00    &         &         & \textbf{Ours. DPQE} & \textcolor[rgb]{ 0,  .69,  .314}{93.24(-0.16)} & \textbf{500.14 } & \textbf{105.14 } \\
    \bottomrule
    \end{tabular}%
    \begin{tablenotes}
    \item[1] \textcolor{orange}{a.} OICSR-GL \cite{qi2021learning}: Structured Sparsity Regularization + Greedy Out-in-Channel P. \textcolor{orange}{b.} PCQAT \cite{b.PCQAT}: Weight Clustering + QAT. \textcolor{orange}{c.} Predictive E+Q \cite{li2023predictive}: Predictive Exit + PTQ. \textcolor{orange}{d.} CEA-MOP \cite{d.CEA-MOP}: NAS + filter P + Coding. \textcolor{orange}{e.} LQP \cite{e.LQP}: Low-Rank Approximation + P + Q. \textcolor{orange}{f.} PD \cite{f.PD}: Weight P via Activation Analysis + D. \textcolor{orange}{g.} KDDP SN \cite{g.KDDPSN}: D + L1 Regularization + P on FC. \textcolor{orange}{h.} HMC \cite{h.HMC}: Low-Rank Tensor Decomposition + Structured P. \textcolor{orange}{i.} DDSL \cite{i.DDSL}: alternating direction method of multiplier based D + P. \textcolor{orange}{j.} Quantized Distill. \cite{j.quantized}: D + Q. \textcolor{orange}{k.} Smart-DNN+ \cite{k.Smart-DNN+}: Q + Bucket Coding. \textcolor{orange}{l.} Hybrid Search \cite{l.hybrid}: NAS + P. \textcolor{orange}{m.} HFPQ \cite{m.hfpq}: channel P + exponential Q. \textcolor{orange}{n.} Deep Compression \cite{han2015deep}: P + weight sharing Q + Huffman Coding. \textcolor{orange}{o.} \cite{wang2020apq} APQ: NAS + P + Q.
    \item[2] Since the accuracy of the original model reproduced in each work is different, both post-compression accuracy and accuracy loss play crucial roles in the comparison. Poor model accuracy reproduction can mitigate the compression accuracy loss, while it cannot conceal a very low post-compression accuracy. Only the coexistence of \textcolor[rgb]{0,.69,.314}{high post-compression accuracy} and \textcolor[rgb]{0,.69,.314}{low accuracy loss} can indicate that a compression achieves \textcolor[rgb]{0,.69,.314}{near-lossless}.
    \item[3] DPQE in this line applies 4w8a (4-bit weight, 8-bit activation) Q for superior post-compression accuracy. DPQE in the line below applies 1w8a Q.
    \end{tablenotes}
  \label{tab:comparison}%
\end{scriptsize}
\end{threeparttable}
\vspace{-3mm}
\end{table*}

Fig. \ref{fig:change} presents the BitOps compression ratios and network inference accuracies resulting from applying each compression technique within the proposed Order of Compression. In the compression process of the three models, each compression technique plays a crucial role in reducing the computation cost of the neural network. A noteworthy observation is the automatic utilization of complementary features among these techniques within the proposed Order of Compression. While one technique may lead to a slight reduction in network accuracy, the following compression techniques may effectively compensate for this loss while continuously decreasing network computation costs. 

Table \ref{tab:comparison} presents a comparison of proposed Order of Compression with other state-of-the-art methods that integrate two or more model compression techniques, such as Distillation, Pruning, Quantization, Early Exit, Coding, Network Architecture Search (NAS), and Low-rank Matrix Factorization. Unlike some of these methods, to maximize generality and hardware friendliness, our chosen compression element is typical and not tailored to a specific network. However, as indicated in the table, the proposed Order of Compression still demonstrates superior performance compared to all other methods. On CIFAR10, for ResNet, the proposed Order of Compression exhibits up to $859\times$ BitOps compression ($4.5\times$ over state-of-the-art predictive E+Q\cite{li2023predictive}); for VGG, the proposed Order of Compression also achieving $105\times$ model size reduction ($1.7\times$ over state-of-the-art LQP\cite{e.LQP}). On CIFAR100, the proposed Order of Compression maintains superior compression performance: $307\times$ BitOps compression ($3.5\times$ over state-of-the-art predictive E+Q\cite{li2023predictive}) and $64\times$ model reduction ($11.5\times$ over state-of-the-art HMC\cite{h.HMC}) for ResNet, and achieves higher post-compression accuracy than others when compressing the same network (5.6\% higher than state-of-the-art HMC\cite{h.HMC}). On difficult dataset ImageNet, proposed Order of Compression still maintains substantial compression ratio and achieve best post-compression accuracy for ResNet (1.10\% higher than state-of-the-art APQ\cite{wang2020apq}).

\section{Conclusion}
\label{sec:conclusion}
To explore interactions and reap the benefits of applying multiple compression techniques, we present a study on the optimal combinational sequence to compress the neural network. Through the proposed roadmap we demonstrate the significant benefits of applying multiple compression approaches and also the importance of the optimal compression sequence. Based on the interactions between two compressions, we establish the optimal combinational sequence law in applying multiple compression techniques. Validated on image-based regression and classification networks across different data sets, our proposed combinational sequence can significantly reduce the computation cost by up to $1000\times$ with ignorable accuracy losses compared to baseline models.
\vspace{5mm}

\bibliography{bio}
\end{document}